\documentclass[10pt,conference,a4paper]{IEEEtran}
\usepackage{times,amsmath,epsfig}
\usepackage{epstopdf}
\usepackage[misc]{ifsym}

\title{Clothing Retrieval with Visual Attention Model}
\author{%
{Zhonghao Wang{\small $~^{1}$}, Yujun Gu{\small $~^{1}$}, Ya Zhang{\small $~^{1}$}\textsuperscript{\Letter}, Jun Zhou{\small $~^{2}$}, Xiao Gu{\small $~^{2}$}}

\vspace{1.6mm}\\
\fontsize{10}{10}\selectfont\itshape
$^{1}$\,Cooperative Medianet Innovation Center, Shanghai Jiao Tong University, Shanghai, China\\
$^{2}$\,Institute of Image Communication and Network Engineering, Shanghai Jiao Tong University, Shanghai, China\\
\fontsize{9}{9}\selectfont\ttfamily\upshape
\{kindredcain,yjgu,ya\_zhang,zhoujun,gugu97\}@sjtu.edu.cn

}
\usepackage{fancyhdr}

\begin{document}
\maketitle

\thispagestyle{fancy}
\fancyhead{}
\lhead{}
\lfoot{978-1-5386-0462-5/17/\$31.00 ~\copyright~2017 IEEE}
\cfoot{}
\rfoot{VCIP 2017, Dec. 10 -- 13, 2017, St Petersburg, U.S.A}

\begin{abstract}
Clothing retrieval is a challenging problem in computer vision. With the advance of Convolutional Neural Networks (CNNs), the accuracy of clothing retrieval has been significantly improved. FashionNet \cite{Liu2016DeepFashion}, a recent study, proposes to employ a set of artificial features in the form of landmarks for clothing retrieval, which are shown to be helpful for retrieval.
However, the landmark detection module is trained with strong supervision which requires considerable efforts to obtain. In this paper, we propose a self-learning Visual Attention Model (VAM) to extract attention maps from clothing images. The VAM is further connected to a global network to form an end-to-end network structure through Impdrop connection which randomly Dropout on the feature maps with the probabilities given by the attention map. Extensive experiments on several widely used benchmark clothing retrieval data sets have demonstrated the promise of the proposed method. We also show that compared to the trivial Product connection, the Impdrop connection makes the network structure more robust when training sets of limited size are used.
\\[1\baselineskip]
\end{abstract}


\section{Introduction}
With the rapid development of e-commerce, product retrieval has become increasingly important for online shopping. The existing keyword based searching strategy often cannot accurately express users' query for specific items. For example, it is difficult to describe the pattern or style of clothes using only keywords. On the other hand, it has become fairly easy for a user to provide images for an item of interests.  As a result, we have witnessed a surge in studies for content-based product retrieval and in particular clothing retrieval \cite{Liu2016DeepFashion}\cite{kiapour2015buy}\cite{xiong2016parameter}. Clothing retrieval is a special type of image retrieval tasks with the following unique characteristics. First, clothes are flexible items and the appearance can be quite different when captured from different shooting angles or worn by different statures. Second, the query images provided by users may be taken in complicated conditions (e.g. street photography) with complex background, diverse shooting angles and lighting conditions, and even with occlusions. On the contrary, most shop images are pictured with clean background, good lighting and positive angle. The differences between shop and consumer images make the clothing retrieval a naturally cross-domain task. As a result, clothing retrieval has always been a challenging problem.

Convolutional Neural Networks (CNNs), which is shown to be quite effective for general image retrieval tasks, have been widely adopted for clothing retrieval \cite{Liu2016DeepFashion}\cite{kiapour2015buy}\cite{xiong2016parameter}.
These retrieval methods produce embedding features vectors through CNNs and search similar images using the distances between these vectors.
The Wh method \cite{kiapour2015buy} learns a similarity measure between different domains for clothing retrieval. The parameter partial-sharing method \cite{xiong2016parameter} adopts CNNs with partial different parameters for different domain images and outperforms the parameter-sharing scheme.

By default, CNN is fed with the entire images. While various data augmentation tricks have been proposed and shown helpful for retrieval, few studies explicitly consider the disturbances in the images such as background and occlusions. To deal with this problem, FashionNet artificially adds features by detecting landmarks from the image and employing the detected landmarks to pool/gate the learned features \cite{Liu2016DeepFashion}. The addition of artificial features improves the performance of deep neural network. However, landmark detection module requires training with strongly supervised landmarks labeled by human, which is both expensive and time-consuming to obtain. 

In this paper, we propose a self-learning Visual Attention Model (VAM) to extracts attention maps from clothing images and develop an attention architecture which connects VAM to neural network. This is an end-to-end model, so that the VAM can learn which locations are worthy of attention automatically during training. We pretrain a fully convolutional network (FCN) \cite{Long2015Fully} to generate attention maps. The attention maps are combined with intermediate feature maps to further generate feature vectors for retrieval. The loss applied to these features trains both the main network and the VAM. To implement the combination, we develop the Impdrop connection inspired by Dropout \cite{Hinton2012Improving} which enhances the robustness of the model.

\begin{figure*}[!htb]
  \centerline{
    \includegraphics[width=\linewidth]{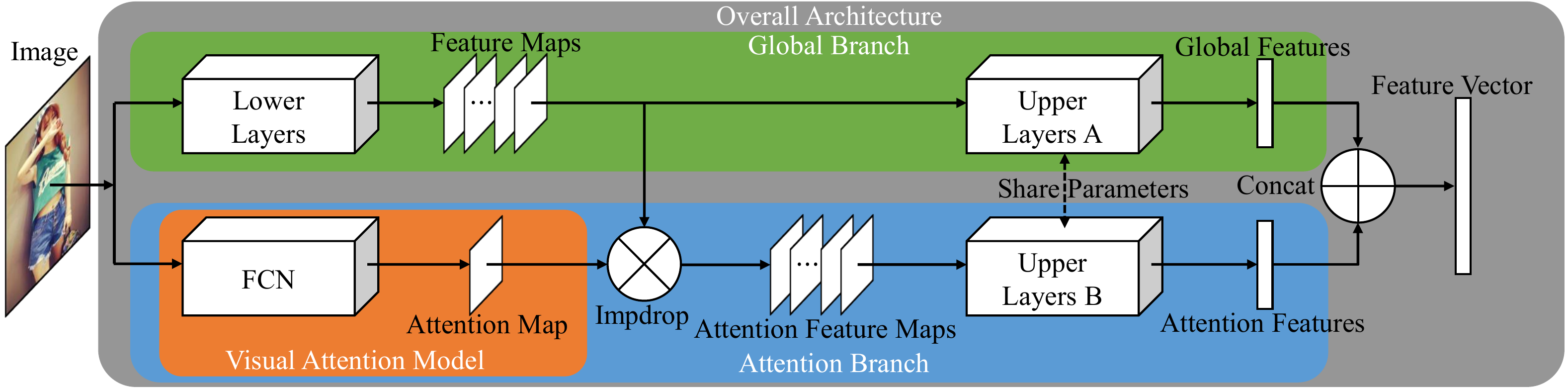}
  }
  \caption{The overall architecture of our model. The architecture consists of: 1) the {\bf Global Branch} produces features maps from lower layers and outputs global features using the upper layers A; 2) the {\bf Attention Branch} extracts the attention map with {\bf VAM}, combines it with the feature maps into attention feature maps using Impdrop and outputs attention features through upper layers B.}\label{fig:architecture}
\end{figure*}

Extensive experiments are performed to validate the effectiveness of the VAM. The experiment results have demonstrated that the proposed attention architecture is effective for different datasets, with different base networks, and under different learning situations. We also experimentally show the advantage of the Impdrop connection over element-wise product connection. Compared to FashionNet \cite{Liu2016DeepFashion}, the current state-of-the-art clothing retrieval method, the proposed VAM-enhanced architecture leads to 24.9\% and 15.9\% improvement in terms of retrieval accuracy at top 20, for the C2S and In-shop data sets \cite{Liu2016DeepFashion}, respectively, with only 1/16 feature length.

The main contributions of our study is as follows. First, we develop a VAM-enhanced architecture for clothing retrieval which can be trained end-to-end and achieve the state-of-art retrieval accuracy. Secondly, we propose the Impdrop connection method which connects the VAM to the main network and helps the architecture to be more robust.

\section{Approach}




The overall architecture of our model is shown in Fig. \ref{fig:architecture}. The whole architecture can be trained end-to-end with triplet loss \cite{schroff2015facenet} and contains two branch networks. The {\em global branch} is based on CNN, and the {\em attention branch} combines the lower layers of CNN with FCN \cite{Long2015Fully} to eliminate the influence of background. The global branch is included to increase the stability of the model, as a single branch with weak attention is not robust enough.

Each image is fed into two branches simultaneously: the lower layers applies low-level feature extraction to the image; while FCN predicts which part of image is more important. The feature maps from the lower layers are fed to both the upper layers of the global branch and `Impdrop' module of the attention branch. The latter combines the feature maps with the attention map from the FCN to reduce the response of unimportant regions (e.g. background) on the feature maps. The output of Impdrop module is called attention feature maps. The attention map, the feature maps, and the attention feature maps have the identical height and width. Besides, the channel number of the feature maps and the attention feature maps are the same. 
The rest part of each branch is defined as upper layers. There are two sets of upper layers which have the same structure and parameters in order to avoid unnecessary increase in parameters. The output of the two branches, global features and attention features, are concatenated to achieve the final feature vector of the image.

\subsection{Visual Attention Model(VAM)}
In many retrieval tasks, the object of interests usually occupies only a part of the image. As a result, background of the image often has non-negligible influence on retrieval results. How to extract features invariant to background is a challenging task. We here propose to leverage the Visual Attention Model (VAM) to partially solve the problem. The VAM generates attention maps of images and reduces the negative influence of background in image retrieval. For clothing retrieval tasks, it is desirable to differentiate clothing regions from the background. Considering the strong co-occurrence relationship between person and clothes, we pre-train a FCN \cite{Long2015Fully} on the datasets of clothes parsing and person segmentation. The values of the output attention map are between 0 and 1, representing the importance of the corresponding regions in the original image.

The VAM is connected to the main network to form an end-to-end network for further training.
As shown in Fig. \ref{fig:architecture}, the attention map is combined with the intermediate feature maps instead of the input image for three main reasons:
\begin{enumerate}
    \item The attention map is forced to have the same height and width as the joined maps (image or feature maps). In CNN the height and width of the feature maps are smaller than the ones of input images, and the task of predicting smaller attention maps is simpler.
    \item Combining the attention maps with the input image causes extra fake edges on the image, which may lead to wrong responses through lower layers. Using the feature maps after low-level feature extraction instead can avoid this influence.
    \item There are overlapping between the receptive fields of the adjacent nodes in the intermediate feature maps. Besides, the intermediate feature maps are thicker in the channel-dimension. Through experiments, we find that these characteristics can increase the stability of the system against the possible error in the attention map.
\end{enumerate}

\subsection{Impdrop Connection}
In combining the attention map and the feature maps, the connection module is designed to force the neural network to focus on the clothing regions so as to eliminate the influence of background. Since the values of the attention map represent the importance of the corresponding regions, one straight-forward solution is to element-wise `multiply' the feature maps with the attention map, named `Product'. With the Product connection, the values of unimportant regions are decreased while the values of important regions are preserved on the feature maps.
\begin{figure}[!htb]
  \centerline{
    \includegraphics[width=3.5in]{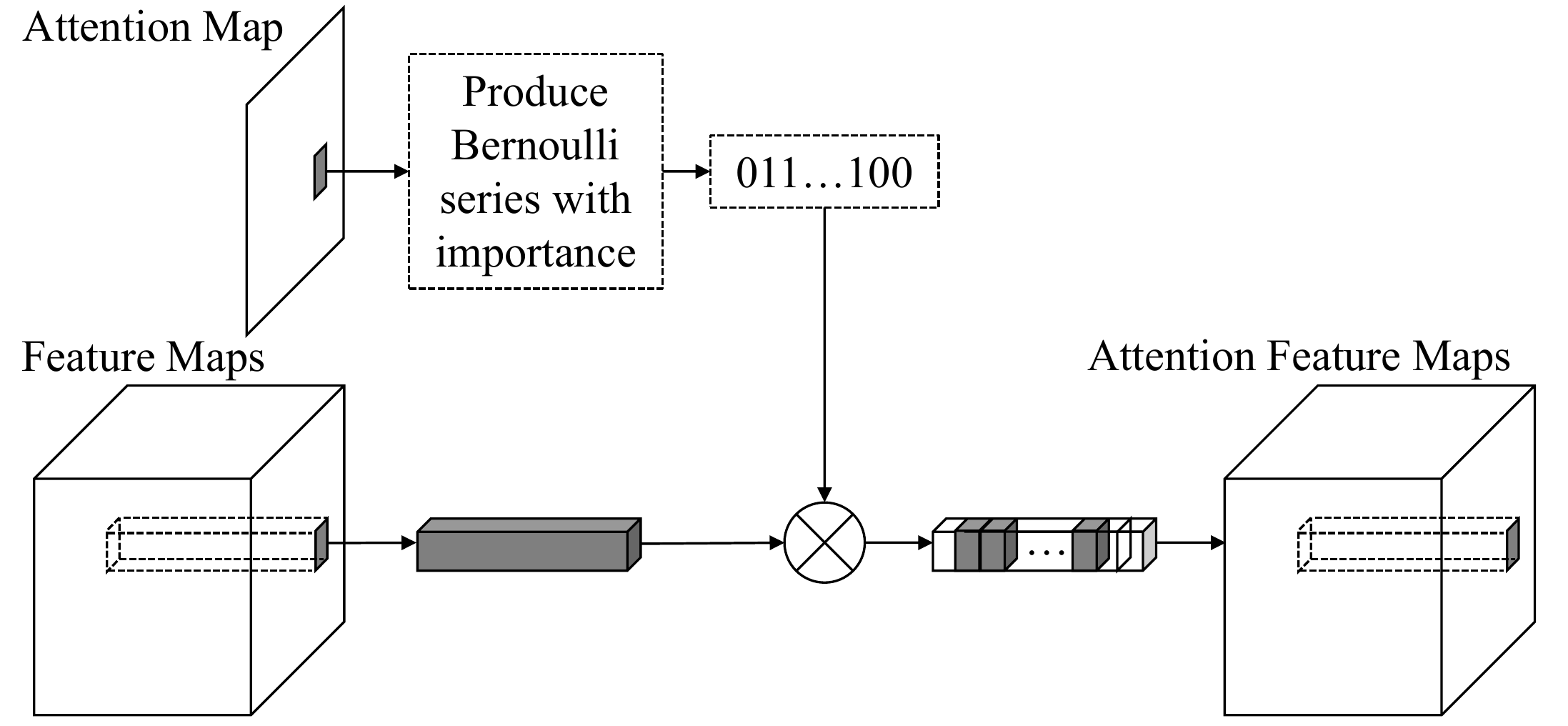}
  }
  \caption{Forward-propagation of Impdrop module in training phase. The heights and widths of the gray parts are $1\times1$.}\label{fig:impdrop1}
\end{figure}

However, the Product method suffers from the problem of overfitting on small datasets.
Inspired by Dropout \cite{Hinton2012Improving}, we design the Impdrop connection which can be seen as the combination of Product and Dropout. In the Impdrop method, each value of the attention map is deemed as the probability that the corresponding region is important. Impdrop thus performs random Dropout on the feature maps using the probabilities given by the attention map in the training phase. As shown in Fig. \ref{fig:impdrop1}, Impdrop generates a Bernoulli series $\textbf{b}_{ij}$ for each value $p_{ij}$ in the attention map, where $i$ and $j$ represent the row index and the column index respectively.
\begin{equation}\label{eq:bernoulli}
\begin{aligned}
P(\textbf{b}_{ij}(c)=k)&=p_{ij}^{(k)}(1-p_{ij})^{(1-k)} \\
&,k=0\ or\ 1, c=1\, ...\, channel
\end{aligned}
\end{equation}
where $c$ is the channel index and $channel$ is the number of channels in the feature maps.

In {\bf forward-propagation}, product operation is applied to the series $\textbf{b}_{ij}$ and the corresponding feature vector $\textbf{x}_{ij}$. The result is written as $\textbf{y}_{ij}$.
\begin{equation}\label{eq:impdropTrainingForward}
\textbf{y}_{ij}= \textbf{x}_{ij}\cdot \textbf{b}_{ij}
\end{equation}

In {\bf backward-propagation}, the gradient $\frac{\partial Loss}{\partial \textbf{x}_{ij}}$ and $\frac{\partial Loss}{\partial p_{ij}}$ are required.
The gradient $\frac{\partial Loss}{\partial \textbf{x}_{ij}}$ is derived similar as the Dropout method.
\begin{equation}\label{eq:impdropTrainingBackward0}
\frac{\partial Loss}{\partial \textbf{x}_{ij}}= \textbf{b}_{ij}\cdot \frac{\partial Loss}{\partial \textbf{y}_{ij}}
\end{equation}
We can calculate the gradient $\frac{\partial Loss}{\partial \textbf{b}_{ij}}$, however, the gradient $\frac{\partial Loss}{\partial p_{ij}}$ cannot be calculated precisely, because $p_{ij}$ is just the probability of $\textbf{b}_{ij}(c)=1$.
Thus, the gradient in the Product method is accepted as an estimation of $\frac{\partial Loss}{\partial p_{ij}}$.
\begin{equation}\label{eq:impdropTrainingBackward1}
\frac{\partial Loss}{\partial p_{ij}}= \sum_{c=1}^{channel} \textbf{x}_{ij}(c)\cdot \frac{\partial Loss}{\partial \textbf{y}_{ij}(c)}
\end{equation}

In the test phase, just like Dropout, Impdrop method abandon randomness. Therefore, the forward-propagation is the same as Product:
\begin{equation}\label{eq:impdropTest}
\textbf{y}_{ij}= p_{ij}\textbf{x}_{ij}
\end{equation}


Compared with Product, Impdrop introduces randomness to the connection between attention maps and feature maps. The addition of such randomness can decrease the risk of overfitting and let the neural network learn more robust features. Consequently, the Impdrop method outperforms the Product method on small training datasets.

\section{Experiments}
\begin{figure*}[!htb]
\begin{minipage}[t]{0.57\linewidth}
\centering
\includegraphics[width=4.08in]{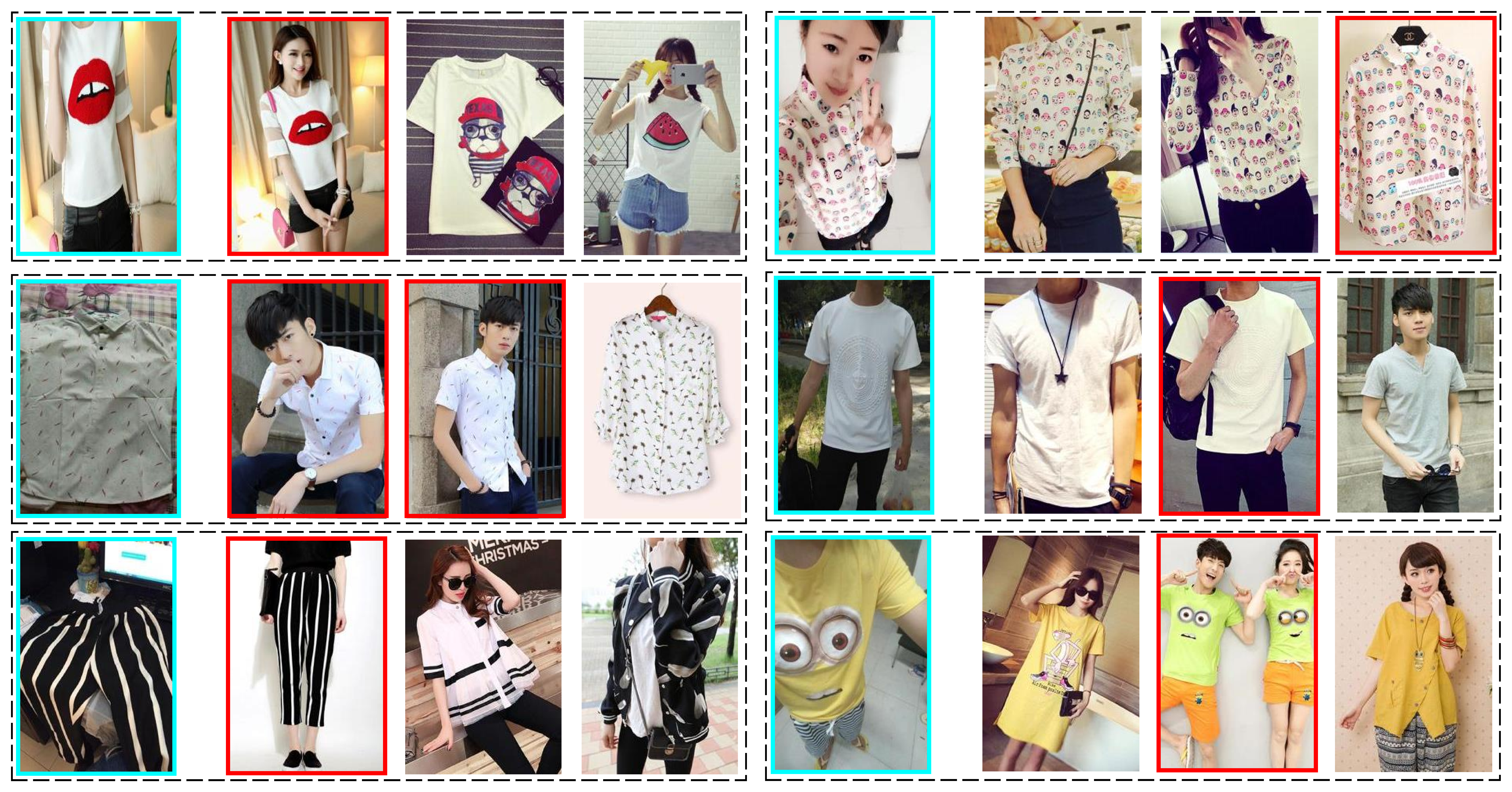}
\footnotesize
(a)Top-3 retrieval examples. Successful matches are marked in red.
\end{minipage}%
\begin{minipage}[t]{0.4\linewidth}
\centering
\includegraphics[width=3in]{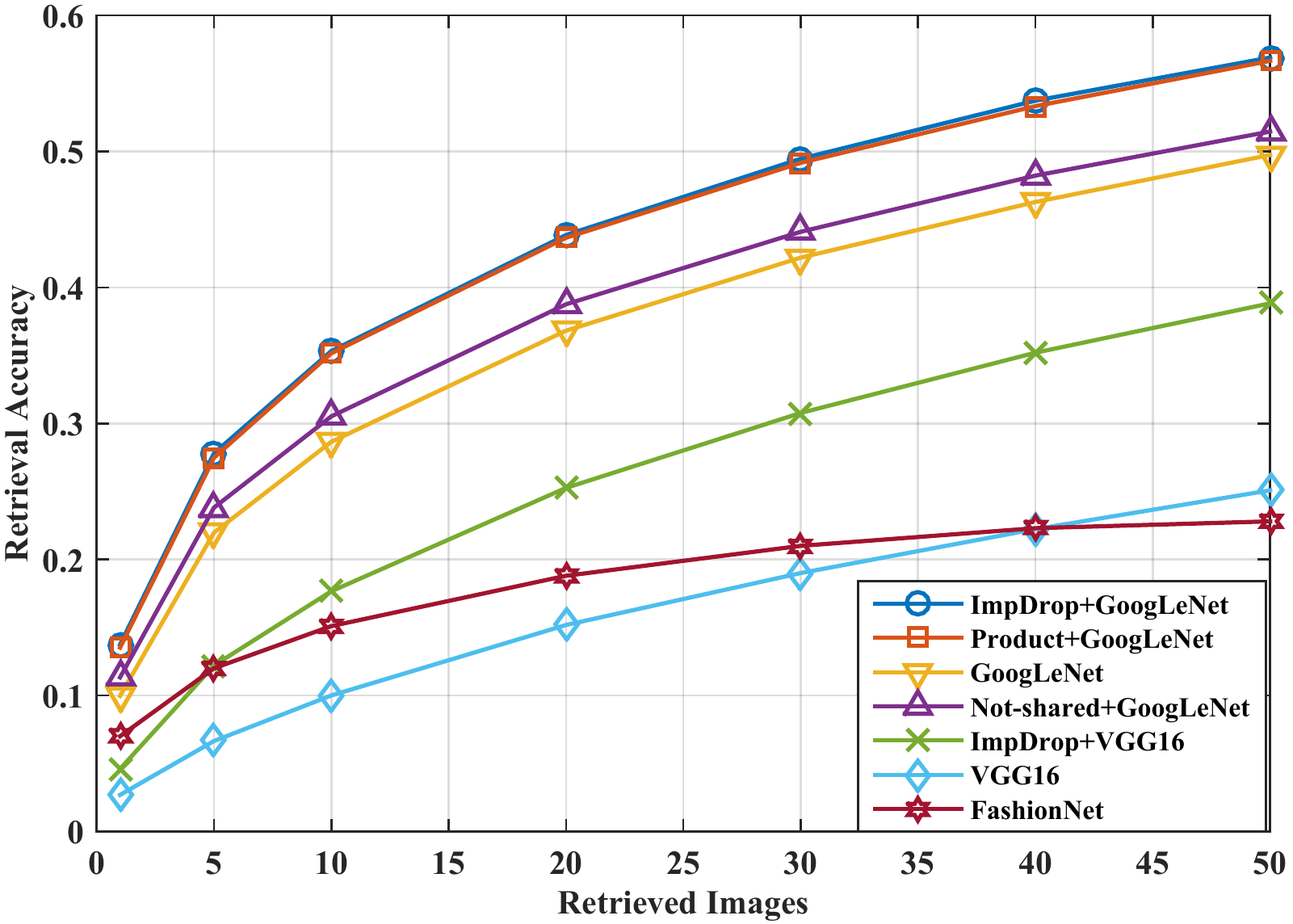}
\footnotesize
(b)Retrieval accuracies of different methods.
\end{minipage}
\caption{Results on DeepFashion consumer-to-shop benchmark \cite{Liu2016DeepFashion}.}\label{fig:result1}
\end{figure*}

\textbf{Data} \ We pre-train our VAM on Fashionista \cite{Yamaguchi2014Paper} based on the model provided by \cite{He2015Deep}. Then, we fine-tune and test our models respectively on DeepFashion \cite{Liu2016DeepFashion} and Street2Shop \cite{kiapour2015buy}. DeepFashion is a clothing dataset which consists of four different kinds of predicting datasets. We only experiment with the two datasets for retrieval. Street2Shop is a clothing dataset for retrieving a street item image in the online shopping images. It has 11 categories of items. We select 3 commonly used categories to do experiments.

\textbf{Triplet generation} \ For cross-domain tasks, we use the positive pairs given by the dataset benchmark and randomly select 40 negative samples for each of them. For in-shop tasks, we randomly choose 100 positive pairs for each class and select random negative samples for them to generate epochs.

\textbf{Evaluation protocol} \ We employ the top-k accuracy to measure the clothing retrieval performance, just as \cite{kiapour2015buy} and \cite{Liu2016DeepFashion}. Items are searched to find the top-k items which are the nearest to the query image according to Euclidean distance. A retrieval task is considered correct only if the top-k items contain the same item as the query image.

\textbf{Implementation details} \ We experiment with the GoogLeNet architecture \cite{Szegedy2015Going} and the VGG-16 architecture \cite{Simonyan2014Very} to generate 256-dim feature vectors, half from each branch. For the VAM, we use the ResNet \cite{He2015Deep} to produce 28x28 attention maps.

\subsection{Results and discussion}\label{sec:results}
In this subsection, ``GoogLeNet'' and ``VGG16'' indicate the architecture shown in Fig. \ref{fig:architecture} without VAM. ``Impdrop+'' and ``Product+'' indicate that VAM is involved with different connection methods. ``Not-shared+'' indicates that upper layers of the global branch and the attention branch do not share parameters. State-of-art methods ``FashionNet'' \cite{Liu2016DeepFashion} and ``Partial-Share'' \cite{xiong2016parameter} are also included in the comparison.

\textbf{DeepFashion} \ We investigate the different methods through experiments on DeepFashion. Fig. \ref{fig:result1} presents the retrieval results on DeepFashion consumer-to-shop dataset. Fig. \ref{fig:result1}(a) illustrates the retrieval results of some typical query images. Fig. \ref{fig:result1}(b) shows the retrieval accuracy of all methods. The ``Not-shared+GoogLeNet'' method increases the network parameters by $98\%$ and gains little improvement compared to ``GoogLeNet''. The ``ImpDrop+GoogLeNet'' and ``Product+GoogLeNet'' methods both gain great improvement compared to ``GoogLeNet" with only $23\%$ parameters more, which can demonstrate the effectiveness of VAM.

To fairly compare with FashionNet \cite{Liu2016DeepFashion}, we also do experiments on the VGG-16 architecture which is used to construct FashionNet. As Fig. \ref{fig:result1}(b) shows, the ``ImpDrop+VGG16'' method achieves higher accuracy than FashionNet in most retrieval tasks, which proves the self-learned VAM is better than the artificial features. Furthermore, the ``VGG16'' and ``ImpDrop+VGG16'' results demonstrate effect of VAM in different architectures. The results of FashionNet are from \cite{Liu2016DeepFashion}.

\begin{table}[!htb]
	\centering
	\caption{Retrieval accuracy for the In-Shop task}\label{tab:inshop}
	\begin{small}
		\begin{tabular}{|c|c|c|c|c|}
			\hline
			Method   &Top-1&Top-5&Top-10&Top-20\\
			\hline
			ImpDrop+GoogLeNet     &\textbf{0.666}&\textbf{0.836}&\textbf{0.887}&\textbf{0.923}\\
			\hline
			GoogLeNet       &0.554&0.758&0.823&0.877\\
			\hline
			FashionNet \cite{Liu2016DeepFashion}       &0.532&0.678&0.725&0.764\\
			\hline
		\end{tabular}
	\end{small}
\end{table}
We also consume the DeepFashion in-shop dataset to investigate the effect of VAM on same-domain retrieval. As shown in Table \ref{tab:inshop}, the VAM still highly improves the performance of GoogLeNet.

\textbf{Street2Shop} \ To investigate the performance of the VAM on different datasets, we experiment with the Street2Shop dataset and the results are shown in Table \ref{tab:results}. The VAM increases the accuracy of the original network on the subsets of all testing categories, demonstrating the effectiveness of VAM. Our method outperforms the state-out-art method ``partial-share'' \cite{xiong2016parameter} in two out of three categories with less parameters.
\begin{table}[!htb]
\centering
    \caption{Retrieval accuracy for the Street2Shop task (top-20)}\label{tab:results}
    \begin{small}
    \begin{tabular}{|c|c|c|c|}
    \hline
    Method & Skirts & Tops & Dresses\\
    \hline
    Impdrop+GoogLeNet & 0.709 & \textbf{0.523} & \textbf{0.621}\\
    \hline
    GoogLeNet & 0.690 & 0.449 & 0.560\\
    \hline
    Partial-Share \cite{xiong2016parameter} & \textbf{0.723} & 0.470 & 0.583\\
    \hline
    \end{tabular}
    \end{small}
\end{table}

\textbf{ImpDrop\&Product} \ Due to the effect of triplet-loss against over-fitting, we train networks on 10\% of the training set using larger triplet loss margin to make the difficult situation artificially. We get the comparable results shown in Table \ref{tab:overfit}. We find that the ImpDrop connection is more robust than the Product connection.
\begin{table}[!htb]
	\centering
	\caption{Retrieval accuracy for the Difficult task}\label{tab:overfit}
	\begin{small}
		\begin{tabular}{|c|c|c|c|c|}
			\hline
			Method   &Top-1&Top-5&Top-10&Top-20\\
			\hline
			ImpDrop+GoogLeNet     &\textbf{0.033}&\textbf{0.087}&\textbf{0.129}&\textbf{0.186}\\
			\hline
			Product+GoogLeNet       &0.022&0.055&0.083&0.127\\
			\hline
            GoogLeNet       &0.023&0.061&0.092&0.138\\
			\hline
		\end{tabular}
	\end{small}
\end{table}

\vfill\eject


\section{Conclusion}
In this paper, we develop an end-to-end VAM-enhanced architecture. The VAM can extract the important positions on the image and enforce the main network concern these positions more to increase the performance. Besides, we invent the Impdrop method which can improve the robustness of the architecture. The experiments demonstrate the effectiveness of our approach which achieves the state-of-art retrieval accuracy.

\section{Acknowledgements}
The work is partially supported by the High Technology Research and Development Program of China 2015AA015801, NSFC 61521062, and STCSM 12DZ2272600.

\bibliographystyle{IEEEtran}

\bibliography{zhwang_VCIP_2017}

\end{document}